\documentclass{article}

\usepackage[preprint]{neurips_2026}

\usepackage[utf8]{inputenc} % allow utf-8 input
\usepackage[T1]{fontenc}    % use 8-bit T1 fonts
\usepackage{hyperref}       % hyperlinks
\usepackage{url}            % simple URL typesetting
\usepackage{booktabs}       % professional-quality tables
\usepackage{amsfonts}       % blackboard math symbols
\usepackage{nicefrac}       % compact symbols for 1/2, etc.
\usepackage{microtype}      % microtypography
\usepackage{xcolor}         % colors
\usepackage{graphicx}

\usepackage{algorithm}
\usepackage{algpseudocode}
\usepackage{amsmath}

\usepackage{booktabs}
\usepackage{tabularx}
\usepackage{multirow}
\usepackage{listings}

\title{Cross-Entropy Games and Frost Training}

\author{%
  Arthur Renard \\
  Xent Labs \\
  \And
  Franck Gabriel \\
  Xent Labs, Université Lyon 1 \\
  \And
  Valentin Hartmann \\
  Xent Labs \\
  \And
  Clément Hongler \thanks{Correspondence to \href{mailto:ch@xentlabs.ai}{\texttt{ch@xentlabs.ai}}} \\
  Xent Labs \\
}

\begin{document}

\maketitle

\begin{abstract}
    We present Frost Training, a method for improving Monte Carlo-based policy optimization for a large family of LLM-as-a-judge tasks called Cross-Entropy Games. The key idea is to exploit the gradient of the reward function in embedding space. This signal is used in the Greedy Coordinate Gradient (GCG) jailbreaking technique; we demonstrate for the first time that it can also be used to boost model training. We validate our method using GRPO training for maximum-likelihood infilling. Frost Training improves the model’s ability to generate high-scoring outputs, reaching higher maximum scores in a best-of-$k$ setting, and does so at an increased speed.
\end{abstract}

\section{Introduction}

%\subsection{Monte Carlo Estimators for Policy Gradient Methods}

\label{subsec:monte-carlo-estimators-for-policy-gradient-methods}

Monte Carlo methods lie at the heart of Reinforcement Learning (RL) for LLMs \cite{schulman2017ppo,shao2024deepseekmath}; most policy optimization methods aimed at training an LLM for a task rely on the following: get an assignment $x$, produce a response $y$, get a reward $R\left(x,y\right)$, and update the model's parameters (``policy'') to shift toward producing higher-scoring outputs \cite{sutton1999policygradient,williams1992reinforce}.

In this paper, we argue that for a natural class of tasks, called Cross-Entropy Games \cite{hongler2025crossentropy,hongler2026cognitive}, we can go beyond the standard Monte Carlo paradigm. The idea is to leverage the implicit differentiable structure of their reward functions to significantly boost a Monte Carlo method. In particular, we show that for max-likelihood infilling tasks, this yields qualitative improvements in the trained model's outputs and reduces training time.

\subsection{Cross-Entropy Games}

Cross-Entropy Games are a family of tasks whose reward function is based on the cross-entropy function of a judge LLM. This space of games generalizes in particular a number of classical tasks: 
\begin{itemize}
\item Max-likelihood infilling \cite{hu2024amortizing}. Given a beginning $x_{<k}$ and an end $z_{<m}$ of a text, the reward for a token string $y_{<L}$ connecting the two is $-\mathrm{CE}\left(y_{<L}z_{<m}\mid x_{<k}\right)=\log\mathbb{P}\left\{ y_{<L}z_{<m}\mid x_{<k}\right\} $, where $\mathrm{CE}$ denotes the cross-entropy loss. 
\item Reverse prompting and jailbreaking \cite{das2024security}. The reverse prompting task consists of finding a prefix $x_{<k}$ to a token string $y_{<L}$ that maximizes the likelihood of the text $-\mathrm{CE}\left(y_{<L}\mid x_{<k}\right)$. Jailbreaking is a special instance of that where $y_{<L}$ is a text like ``Of course I will help you with that.'' 
\item Reinforcement Learning as a Pretraining Objective (RLP) \cite{hatamizadeh2026rlp}. RLP trains models to reason about the next token of a prefix by producing a chain of thought that minimizes that token's cross-entropy: the reward is $-\mathrm{CE}\left(x_{k-1}\mid x_{<k-1}y_{<L}\right)$, where $y_{<L}$ is the chain of thought. 
\end{itemize}
In Section~\ref{ref:differentiable_structure}, we discuss a key feature of Cross-Entropy Games: their implicit differentiable structure.

\subsection{Differentiable Structure}

\label{ref:differentiable_structure}

A central feature of the cross-entropy function of an LLM is its implicit differentiable structure. For instance, let $x_{<k}$ and $z_{<m}$ be two sequences of tokens and $y$ be a token, then 
\[
\mathrm{CE}\left(yz_{<m}|x_{<k}\right)=\mathrm{CE}\left(z_{<m}|x_{<k}y\right)-\log\mathbb{P}\left\{ y|x_{<k}\right\} ,
\] where $\mathrm{CE}\left(z_{<m}|x_{<k}y\right)$ can be written as the composition of the one-hot embedding function $y\mapsto e_{y}\in\mathbb{R}^{V}$ with a smooth function $\mathcal{S}_{x_{<k}}^{z_{<m}}:\mathbb{R}^{V}\to\mathbb{R}$ (corresponding to the processing performed by the LLM after the embedding, composed with the cross-entropy loss function).

This feature enables us to extract an important signal from the reward function: we can use the gradient of the reward with respect to the one-hot embedding space to estimate not only the score of the move $y$, but also of the alternate values $\tilde{y}\in \mathcal{V}$: 
\begin{equation}
\left(\mathrm{CE}\left(yz_{<m}|x_{<k}\right)\right)_{\tilde{y}\in \mathcal{V}}
\approx\left(\mathrm{CE}\left(z_{<m}|x_{<k}y\right)+\nabla\mathcal{S}_{x_{<k}}^{z_{<m}}\left(e_{y}\right)\cdot\left(e_{\tilde{y}}-e_{y}\right)-\log\mathbb{P}\left\{ \tilde{y}|x_{<k}\right\} \right)_{\tilde{y}\in \mathcal{V}}.\label{eq:taylor-approx}
\end{equation}
Note that this estimation of all the $\tilde{y}\in\mathcal{V}$ possibilities can be computed in a \emph{single} forward-backward pass (the $\log\mathbb{P}$ is the direct result of the model's output given $x_{<k}$). This can be generalized to multi-token settings, and more general Cross-Entropy Game rewards, as in our experiments.

\subsection{Contributions}

The key contribution of this work is to demonstrate the exploitability of the gradient signal above to improve the training of LLMs on Cross-Entropy Games, enabling them to output higher-scoring moves faster. In particular, we show how this idea can be implemented to improve a training algorithm such as GRPO, yielding Frost-GRPO; we demonstrate the power of this method on a specific Cross-Entropy Game (maximum likelihood infilling task) and perform a number of analyses to validate the design choices and demonstrate the effectiveness of Frost-GRPO in this setting. 

\subsection{Related Work}

\paragraph{Monte Carlo Policy-Gradient Methods for LLMs.}

Policy-gradient methods such as REINFORCE \citep{williams1992reinforce} and PPO~\citep{schulman2017ppo} are standard starting points for Reinforcement Learning with language models. Recent variants, such as GRPO \citep{shao2024deepseekmath}, compare multiple on-policy completions for the same prompt using a group-relative baseline. Frost Training is complementary to these methods: rather than changing the policy-gradient objective directly, it modifies the set of samples used in the update using first-order information about the reward.

\paragraph{Gradient-Guided Token Search and Inference.}

Several methods use gradients with respect to token representations to guide discrete search and inference. HotFlip \citep{ebrahimi2018hotflip} uses one-hot gradients to rank token substitutions, AutoPrompt \citep{shin2020autoprompt} constructs prompts, and GCG \citep{zou2023universal} optimizes adversarial universal suffixes. Related approaches also use one-hot gradients of smooth constraints to steer decoding at inference time \citep{kumar2022gradient}, including with Langevin-based energy minimization \citep{qin2022cold}. We use a similar local gradient signal, but apply it during training rather than inference: gradients propose candidate edits, while the accepted samples are then used in a policy-gradient update.

\paragraph{Reward Gradients for Training.}

In continuous optimization, several methods combine sampling with first-order reward information. For example, within the Cross-Entropy Method framework, \citep{bharadhwaj2020gradcem,huang2021cemgd} interleave CEM sampling with gradient optimization of actions. Frost differs in that it operates in discrete token space: gradients are only used to propose local edits, subsequently filtered by exact reward evaluation.

In the context of training generative adversarial networks, TaylorGAN \citep{lin2020taylorgan} exploits the differentiability of the discriminator to reduce the variance of REINFORCE. It uses the Taylor approximation of the reward to average the reward locally over neighborhoods of sampled sequences. TaylorGAN interpolates between pure REINFORCE, which is unbiased but high-variance, and straight-through estimators, which have lower variance but higher bias. Frost Training uses first-order reward information differently. Unlike TaylorGAN, the Taylor approximation is used only to propose local mutations, not to assign their reward. The resulting update is still biased, since the accepted samples are not drawn from the current policy distribution. However, only mutations that improve their original sample are incorporated. Thus, the bias mostly affects the proposal distribution toward higher-scoring samples, while the reward signal used for selection and the advantage computations are obtained from exact reward evaluations. 
%Rather than using the Taylor approximation to assign or locally average rewards, it uses the approximation only to  propose candidate token
%substitutions. These candidates are then evaluated under the true reward, and only improving mutations are included in the GRPO update. 
%The resulting update is still biased, since the accepted samples are not drawn from the current policy distribution. However, unlike TaylorGAN, the Taylor approximation is used only to propose local mutations, not to assign their reward. Moreover, only mutations that improve their original
%sample are incorporated. Thus, the bias affects only the proposal distribution toward higher-scoring samples, while the reward signal used for selection, advantages, and gradient step remains exact. 

\section{Frost Training}

\label{sec:frost-training}

We introduce \textit{Frost Training}: alongside the moves the model samples during training, we also score \textit{alternate moves}, that is, moves the model did not sample on its own (hence the name, after Robert Frost's poem \textit{The Road Not Taken}). We identify them via a first-order Taylor expansion of the scoring function in token-embedding space (Section \ref{ref:differentiable_structure}), which yields an approximated score for every one-token mutation of every sampled move at the cost of a single backward pass through the scoring function. The mutations whose exact score beats the original sample are fed back into the training signal.

As a proof of concept, we fine-tune an LLM $\pi_{\theta}$ on a Cross-Entropy Game. We apply the Frost Training idea to the GRPO training algorithm \cite{shao2024deepseekmath}, yielding \textit{Frost-GRPO}, which we compare to regular GRPO on a specific Cross-Entropy Game (an infilling task, see Section \ref{subsec:frost-training-setup}). 

\subsection{Frost-GRPO Algorithm}

\label{sec:frost-algorithm}

We now introduce Frost-GRPO, which is defined as a modification of the classical GRPO algorithm. Given a prompt $P$ (typically for max-likelihood infilling, $P$ will contain the given beginning and end of the task), GRPO samples $K$ on-policy moves $\{y_{k}\}_{k=1}^{K}\sim\pi_{\theta}(\cdot\mid P)$, scores them (using the exact reward function), and updates $\theta$ by using the group advantages \cite{shao2024deepseekmath}. A Frost-GRPO step follows the same procedure, but with one extra \emph{mutation step} inserted between the on-policy sample and the gradient step:
\begin{itemize}
\item For each sampled move (the `parent'), we compute \emph{approximated scores} for all possible mutations (the `children') using a Taylor approximation of the reward.
\item \emph{Proposal}: We select a number of the children (across all parents) with the highest approximated scores, and compute their exact scores.
\item \emph{Replacement}: If a parent has a child with a higher exact score than itself, it gets replaced by the highest-scoring of its children. 
\end{itemize}
Algorithm \ref{alg:frost-grpo} summarizes the procedure; the two stages (proposal and replacement) are detailed below.

\paragraph{Frost Proposal}

For each parent move $y_{k}$, position $j<L$, and replacement token $v\in\mathcal{V}$, write $y_{k}^{j\to v}$ for the \emph{candidate move} obtained by setting the token at location $j$ to be $v$. We approximate $R(y_{k}^{j\to v})$ by a first-order Taylor approximation (as in (\ref{eq:taylor-approx}) above):
\begin{equation}
a(k,j,v):=R(y_{k})+\langle e_{v}-e_{y_{k}[j]},\ \nabla_{e_{y_{k}[j]}}R(y_{k})\rangle.\label{eq:approx-scores}
\end{equation}

The gradient is obtained from the reward, so all $K\cdot L\cdot|\mathcal{V}|$ approximated scores come out of a single backward pass. We apply gating, keeping only candidates with $\pi_{\theta}(v\mid P,y_{k}^{<j})>\tau$. The gating caps the magnitude of $\log\pi_{\theta}(v\mid P,y_{k}^{<j})$ of any accepted off-policy candidate that can contribute to the REINFORCE surrogate, preventing a single low-probability acceptance from dominating the policy gradient. Then the $D$ candidates with the largest approximated value (among all the gated candidates) form the \emph{Frost candidate group}.

\paragraph{Frost Replacement}

From the parent moves $y_{k}$ for $k=1,\ldots,K$ and the $D$ Frost candidates $\tilde{y}_{d}$ for $d=1,\ldots,D$, a Frost group of $K$ moves is computed as follows:
\begin{itemize}
\item We compute the exact scores $\tilde{r}_{d}$ of the $D$ Frost candidates for $d=1,\ldots,D$.
\item The Frost group $y_{k}^{*}$ for $k=1,\ldots,K$ is formed as follows: for each $k=1,\ldots,K$, take $y_{k}^{*}$ to be the highest-scoring Frost candidate that descends from $y_{k}$ if its exact score is higher than that of $y_{k}$, and to be $y_{k}$ if no such candidate exists. 
\end{itemize} 
Note that the policy gradient estimator based on the Frost group is biased, as the accepted mutations in the Frost group are not drawn from $\pi_{\theta}$. However, every accepted candidate is a verified strict improvement over an on-policy parent under the exact reward $R$.

% ---------------------------------------------------------------

\begin{algorithm}[t]
\caption{Frost-GRPO step. Shown for a single prompt $P$; the batch dimension is processed in parallel.}
\label{alg:frost-grpo}

\begin{algorithmic}[1]
\Require Player $\pi_{\theta}$, reward $R:\mathcal{V}^{L}\to\mathbb{R}$, prompt $P$, group size $K$, move length $L$, discovery budget $D$, probability gate $\tau$, KL coefficient $\beta$.

\State Sample $K$ on-policy moves $\{y_{k}\}_{k=1}^{K}\overset{\mathrm{iid}}{\sim}\pi_{\theta}(\cdot\mid P)$ and save the values $\pi_{\theta}(v\mid P,y_{k}^{<j})$ for all $k=1,\ldots,K$, $j=1,\ldots,L$, and $v\in\mathcal{V}$.\hfill{}\Comment{player forward}

\State Score them: $r_{k}\gets R(y_{k})$. \hfill{}\Comment{reward forward}

\State Compute the Taylor approximation $a(k,j,v)$ of $R(y_{k}^{j\to v})$ via Eq. (\ref{eq:approx-scores}) for all $k,j,v$. \hfill{}\Comment{reward backward}

\State Among $(k,j,v)$ with $v\neq y_{k}[j]$ and $\pi_{\theta}(v\mid P,y_{k}^{<j})>\tau$, take the $D$ candidates $(k_{d},j_{d},v_{d})_{d=1}^{D}$ with the highest $a$-value.

\State Compute the exact scores of the $D$ candidates: $\tilde{y}_{d}\gets y_{k_{d}}^{j_{d}\to v_{d}}$, $\tilde{r}_{d}\gets R(\tilde{y}_{d})$. \hfill{}\Comment{reward forward}

\State For each parent \(y_k\), let
\[
(y_k^*, r_k^*) =
\begin{cases}
    (\tilde{y}_d^*, \tilde{r}_d^*) & \text{if the highest-scoring } \tilde{y}_d^* \text{ with } k_d = k \text{ satisfies } \tilde{r}_d^* > r_k, \\
    (y_k, r_k) & \text{otherwise.}
\end{cases}
\]

\State Compute the GRPO advantage $A_k \leftarrow r_k^* - \frac{1}{K}\sum_{k'=1}^K r_{k'}^*$.

\State Update $\theta$ with the GRPO loss on $\{(y_k^*, A_k)\}_{k=1}^K$ and KL coefficient $\beta$.

\end{algorithmic}
\end{algorithm}

\section{Frost Training for Infilling Games: Experimental Results}

\label{sec:experiments}

In this section, we demonstrate experimentally the main thesis of this paper: we can leverage the gradient coming from the differentiable structure of Cross-Entropy Games to boost model training via Frost-GRPO. Our experiments are performed on an infilling task and demonstrate a substantial boost over vanilla GRPO training. 

\subsection{Task and Model Setup}\label{subsec:frost-training-setup}
\begin{itemize}
\item Given a source document, we extract the first $k=8$ tokens as the beginning $x_{<k}$ and an $m=64$-token end $z_{<m}$ that starts $24$ tokens after $x_{<k}$ in the document.
\item The source documents are taken from stories of the Cosmopedia dataset \cite{huggingface_cosmopedia}. The first $128$ stories form the validation set; the remaining $\sim 10^{6}$ ones form the training pool.
\item The player conditions on a chat-formatted prompt $P$ that contains both $x_{<k}$ and $z_{<m}$, and produces an $L$-token move $y\in\mathcal{V}^{L}$. 
\item The judge $\pi_{J}$ scores the move using the log-likelihood (negative cross-entropy loss):
\[
R(y):=\log\pi_{J}(y\,z_{<m}\mid x_{<k})=-\mathrm{CE}\left(y_{<L}z_{<m}|x_{<k}\right).
\]
\item Our experiments focus on $L=8$ (in addition, we sweep over $L=4,12$ in some experiments), which is shorter than the original gap (thus requiring the model to output a text different from the original). 
\item All experiments use Qwen3-14B (\texttt{BF16}) \cite{qwen3technicalreport} for both the judge $\pi_{J}$ and the player $\pi_{\theta}$. 
\item We fine-tune the player $\pi_{\theta}$ with a LoRA adapter on every attention and MLP projection (rank and $\alpha$ both 256, no dropout). The judge $\pi_{J}$ and the KL reference $\pi_{\mathrm{ref}}$ are both the (frozen) Qwen3-14B model.
\end{itemize}

\subsection{Approximation Setup}

The score approximation has two terms: the first one (which is the GCG-inspired term), coming from the differentiation of the LLM with respect to its input, and the second coming from the value of the output distribution at the sampled token (which is a different signal, which regularizes the output). In our experiments, as a means to show the value of the gradient signal, we only use the first term to compute the approximated scores.

\subsection{Selection Rule Experiment}\label{subsec:selection-rule-experiment}

Algorithm \ref{alg:frost-grpo} relies on a selection rule to pick the $D$ Frost candidates out of the $K\cdot L\cdot|\mathcal{V}|$ possible single-token mutations. We first investigate numerically the choice of that selection rule on a fixed checkpoint (without any training). This corresponds to the \emph{first Frost-GRPO step} performed before training starts, in order to answer the following:
\begin{itemize}
\item Does the GCG-inspired term carry a non-trivial signal (i.e. does it outperform other selection rules)?
\item Does the probability gate matter at the discovery budgets we actually train with?
\end{itemize}

\paragraph{Four Selection Rules}

We study four selection rules:
\begin{itemize}
\item \textbf{Random:} pick $D$ candidates uniformly at random. No ranking signal. 
\item \textbf{TopProb:} pick the $D$ candidates with the largest $\pi_{\theta}$. Uses the player's prior; ignores the judge. 
\item \textbf{Taylor}: pick the $D$ candidates with the largest reward approximation $a$, without a probability filter. Uses the judge's gradient signal; ignores the player's prior. 
\item \textbf{Taylor-Gated}: the rule of Algorithm \ref{alg:frost-grpo}: top-$D$ by $a$ among candidates with probability at least $\tau=10^{-4}$. 
\end{itemize}
These rules expose three independent gaps. \textbf{Random vs. TopProb} measures whether the player prior contains any usable signal. \textbf{TopProb vs. Taylor} measures the value of the judge's first-order signal relative to the policy prior. \textbf{Taylor vs. Taylor-Gated} measures the value of suppressing low-probability tokens from the Taylor ranking.

\paragraph{Protocol}

For each of the $128$ held-out validation prompts we sample $K$ on-policy moves, compute $a$ and $\pi_{\theta}$ across the full $K\cdot L\cdot|\mathcal{V}|$ grid once, and let each rule pick its top $D$. We then exact-score the union of selected mutations under the judge and run the Frost replacement step (with the chosen selection rule) of Algorithm \ref{alg:frost-grpo}. The same pool of approximated scores is re-used across all four rules, so any difference in outcome is attributable to the ranking rule alone.

We track three per-prompt metrics, averaged across the $128$ prompts with $\pm1$ standard error bands.

\begin{itemize}
    \item \textbf{Best-of-$K$ post-replacement.} The score of the single best move after Frost replacement. This is the metric that matters for training since GRPO updates against the group-mean baseline and benefits from a high maximal score. 
    \item \textbf{Hit rate.} Fraction of selected mutations that strictly outscore their parent. 
    \item \textbf{Mean lift.} Mean of $\tilde{r}_{d}-r_{k}$ over the accepted mutations (those that strictly outscore their parent). Captures the size of an average successful bet. 
\end{itemize}

The no-replacement baseline (best-of-$K$ before any mutation, $\max_{k}r_{k}$) is shown as a horizontal reference. The fraction of parents replaced by each rule is reported in Appendix \ref{app:frac-replaced}.

\paragraph{Findings}

\begin{figure}[H]
  \centering
  \includegraphics[width=0.85\textwidth]{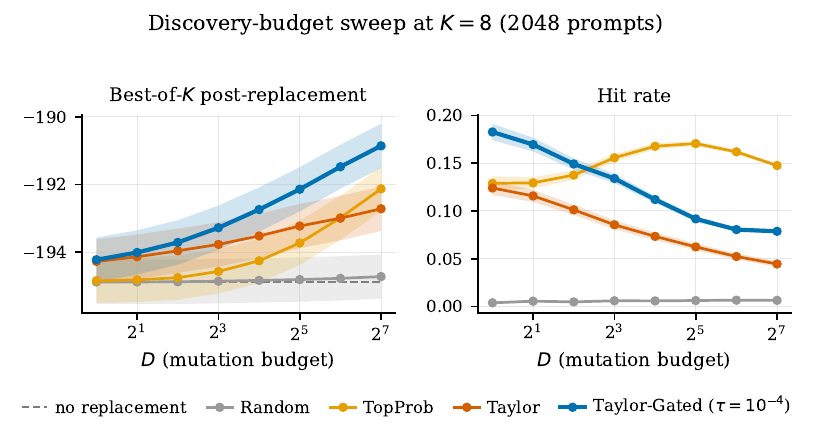}
  \caption{Discovery diagnostics for the four selection rules at $K=8$ as the discovery budget $D$ ranges over $1,2,4,8,16,32,64,128$, averaged over $128$ validation prompts. Left: best-of-$K$ post-replacement (primary metric, with the no-replacement baseline as a horizontal reference). Right: hit rate.}
  \label{fig:sweepD}
\end{figure}

\begin{figure}[H]
  \centering
  \includegraphics[width=0.9\textwidth]{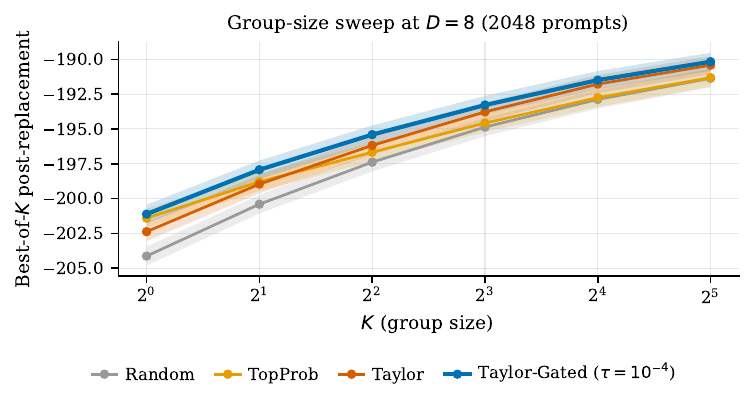}
  \caption{Best-of-$K$ post-replacement for the four selection rules with $K\in\{1,2,4,8,16,32\}$ at fixed $D=8$, averaged over $128$ validation prompts per $K$.}
  \label{fig:sweepK}
\end{figure}

We sweep over $D\in\{1,2,4,8,16,32,64,128\}$ at $K=8$ (Figure \ref{fig:sweepD}), and separately sweep over $K\in\{1,2,4,8,16,32\}$ at $D=8$ (Figure
\ref{fig:sweepK}). 

Taylor-Gated dominates on best-of-$K$ at every $D$ (Figure \ref{fig:sweepD}, left). The gap to the next-best rule grows from roughly $+1$ nat at $D=1$ to roughly $+2$ nats at $D=128$. Random sits at the no-replacement baseline, as it should: uniform sampling carries no signal.

The ordering is robust across group sizes (Figure \ref{fig:sweepK}). Taylor-Gated stays on top of best-of-$K$ at every $K$ tested, with the four rules separating in the same order as in Figure \ref{fig:sweepD}. Absolute best-of-$K$ improves with $K$ (a larger group has a stronger natural max), and Taylor-Gated's lift over the next-best rule is essentially constant across this range. The training group sizes $K\in\{4,8,16\}$ all sit inside this regime.

\subsection{Training Protocol}
\begin{itemize}
\item \textbf{Optimizer:} We use AdamW with learning rate $10^{-7}$ and PyTorch defaults for the remaining hyperparameters ($\beta_{1}=0.9$, $\beta_{2}=0.999$, $\epsilon=10^{-8}$, weight decay $10^{-2}$). KL coefficient $\beta=0.1$. Sampling is on-policy at temperature $1.0$. Each training step uses a batch of $4$ texts.

\item \textbf{GRPO implementation:} For each item $(x_{<k},z_{<m})$ in a batch of $4$ and a corresponding prompt $P=P\left(x_{<k},z_{<m}\right)$, we sample $K$ on-policy moves from $\pi_{\theta}(\cdot\mid P)$ and score each one exactly under the judge. The advantage is $A_{k}=r_{k}-\tfrac{1}{K}\sum_{k'}r_{k'}$, against the group mean. The REINFORCE surrogate is $-\tfrac{1}{K}\sum_{k}A_{k}\log\pi_{\theta}(y_{k}\mid x_{<k},z_{<m})$, with no log-ratio clipping. The KL penalty is the per-position full-vocabulary $\mathrm{KL}(\pi_{\theta}\,\|\,\pi_{\mathrm{ref}})$ summed over move positions and averaged over the group, weighted by $\beta$. Frost uses the same loss, applied to the post-replacement group of size
$K$.
    \item \textbf{Methods and matched-compute pairing:} We compare GRPO (no replacement: $K$ on-policy moves, $K$ judge forwards per step) and Frost (Algorithm \ref{alg:frost-grpo} with $D=K$, Taylor-Gated selection at $\tau=10^{-4}$, strict-improvement acceptance: $K$ on-policy moves plus $D$ Frost candidates, $2K$ judge forward passes plus one judge backward pass through the embedding layer). Two pairs are studied at matched judge-forward-pass count per step:
    \begin{itemize}
        \item \textbf{Canonical pair:} Frost $K=4$ versus GRPO $K=8$ ($8$ judge forwards per step). 
        \item \textbf{Larger-budget pair:} Frost $K=8$ versus GRPO $K=16$ ($16$ judge forwards per step). 
    \end{itemize}

    % We sample for Frost Training half as many on-policy moves per step and run the player backward over a group of size $K$ rather than $2K$, against a single extra judge backward pass whose cost is dominated by the embedding-layer matrix multiply. We do not separately match wall-clock time.

    Per training step at group size K, GRPO performs one player forward and one player backward over a group of size K, one judge forward over a group of size K, and zero judge backwards. Frost-GRPO at the same group size K performs one player forward and one player backward over a group of size K, two judge forwards (one over the K parents and one over the D = K candidates), and one judge backward. The judge backward is cheap because its cost is dominated by the embedding-layer matrix multiply. We do not separately match wall-clock time.

    \item \textbf{Validation:} Every $50$ training steps we sample $8$ on-policy moves per validation story under the current player and exact-score them under the judge. We track four metrics:

    \begin{itemize}
        \item \textbf{Mean reward}: average exact reward across the $8$ samples;
        \item \textbf{Best-of-8}: best exact reward across the $8$ samples;
        \item \textbf{Score variance:} per-story variance of the exact rewards across the $8$ samples;
        \item \textbf{Token entropy}: per-token entropy of the player's distribution at the sampled moves, averaged over positions and stories.
    \end{itemize}

\end{itemize}

\subsection{Canonical Training Experiment Results}

\begin{figure}[H]
\centering \includegraphics[width=1\linewidth]{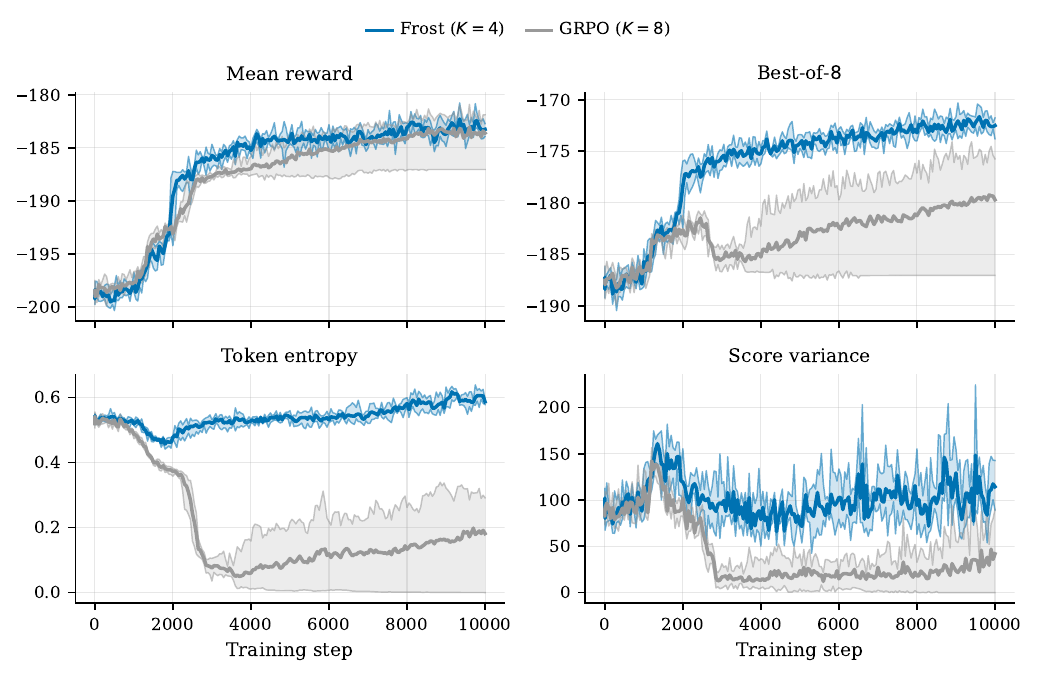}
\caption{Validation curves over training steps for Frost ($K=4$) and GRPO ($K=8$) at the canonical compute-matched configuration ($L=8$, $8$ judge forwards per step). $2\times2$ grid: mean reward, best-of-$K$, token entropy, score variance. Solid lines show the per-step mean across 5 seeds and the shaded band spans the per-step min–max envelope.}
\label{fig:canonical}
\end{figure}

In Figure \ref{fig:canonical}, we compare Frost ($K=4$, $D=4$) and GRPO ($K=8$) over five independent training seeds in the canonical matched-forward setting at $L=8$. Both methods cost $8$ exact judge forwards per step.

Key observations are the following:
\begin{itemize}
    \item The validation best-of-8 trajectories separate early across seeds, and Frost remains consistently ahead throughout training. GRPO exhibits substantially larger seed-to-seed variation: some runs improve late, but all the trajectories remain below Frost throughout.
    \item Frost is also more stable across seeds. Its mean-reward and best-of-8 curves are tightly clustered while GRPO has a wider envelope, especially on best-of-8.
    \item GRPO’s token entropy and score variance tend to drop sharply after the early phase, while Frost maintains higher entropy and substantially higher score variance. This is consistent with GRPO concentrating probability mass on a narrower set of completions.
    \item On the mean-reward panel the two methods finish close, with GRPO slightly ahead in some cases. Across seeds, the mean-reward curves are relatively close, while best-of-8, entropy, and score variance separate them more clearly.
\end{itemize}

Overall, Frost’s advantage is not that it simply raises the average reward, but that it more reliably produces high-scoring samples while preserving diversity.

\subsection{Robustness with Respect to Move Length and Group Size}

\label{sec:scaling-move-length}

\begin{figure}[H]
\centering \includegraphics[width=1\linewidth]{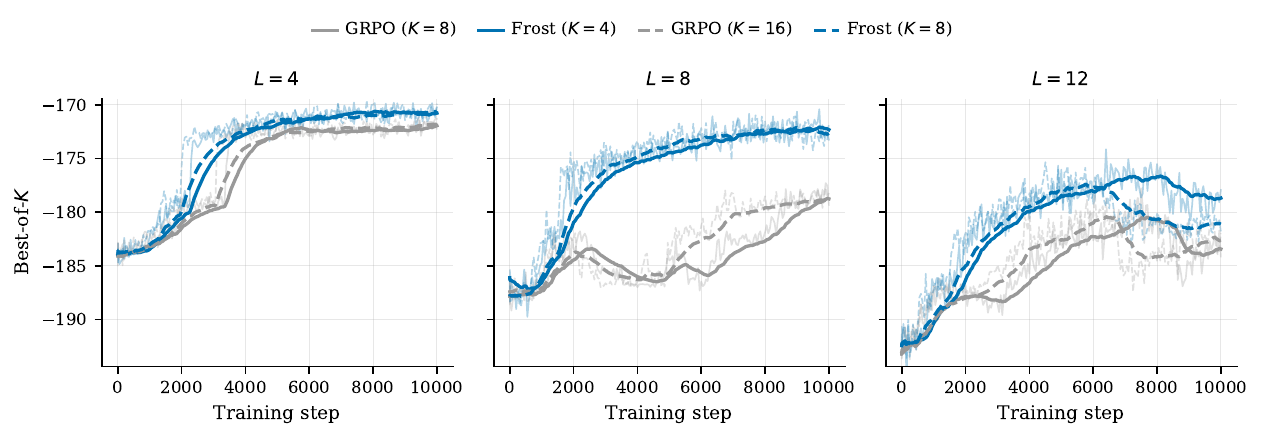}
\caption{Best-of-$K$ over training step for $L\in\{4,8,12\}$. Each panel shows the four matched-compute curves: GRPO $K=8$, Frost $K=4$ (canonical pair, $8$ judge forwards per step), GRPO $K=16$, Frost $K=8$ (larger-budget pair, $16$ judge forwards per step). We apply smoothing over the training steps to generate the solid line.}
\label{fig:l-sweep} 
\end{figure}

In this subsection, we present two variations of the canonical experiment above: we sweep over move size $L\in\left\{ 4,8,12\right\} $ and the two group size configurations described above (Figure \ref{fig:l-sweep}). We find that the Frost-GRPO advantage is consistent across these parameters; at the same time, doubling the GRPO group size or the Frost-GRPO group size does not drastically improve the performance.

Looking at the results from a computational efficiency point of view, we find the following: with $D=K$ Frost-GRPO candidates per step, one Frost-GRPO step at group size $K$ matches or beats the best-of-$K$ trajectory of GRPO at group size $2K$ (which is the GRPO setup that is equivalent in the number of forward passes). Frost trades one judge backward pass for an \emph{effective doubling of the on-policy group}. 

\subsection{Sample Analyses}
\label{sec:sample-analyses}

Inspecting the sampled outputs in the canonical setup reveals an interesting qualitative difference between GRPO and Frost-GRPO: the samples from GRPO go through a ``local optimum'' phase that in many cases repeats the beginning of the text to be infilled, eventually exiting that phase to produce outputs that are also consistent with the end of the text. At the same time, Frost-GRPO climbs much faster toward content consistent with both the beginning and the end of the text, avoiding such a phase.

\subsection{Limitations}\label{sec:limitations}

While the experiments on Frost-GRPO clearly show the presence of a signal in a natural setup, it would be desirable to further validate our findings experimentally by studying a number of directions:
\begin{itemize}
\item GCG approximation: going beyond the GCG approximation, in particular adding the second term of Eq. (\ref{eq:taylor-approx}) to investigate its regularization effect on training;
\item Other training algorithms: validating the advantage of Frost over variants of GRPO, or different training hyperparameters;
\item Other models: validating the signal in the gradient for larger models, or MoE models;
\item Other games: gradient signals exist for all Cross-Entropy Games (including some where the judge and the player are simultaneously trained). 
\end{itemize}

\section{Conclusion}

\label{sec:conclusion}

In this article, we argue that tasks with differentiable rewards, and Cross-Entropy Games in particular, contain a reward gradient signal that standard Monte Carlo policy optimization does not take into account. 

This signal can be exploited to boost training: Frost-GRPO uses it to propose alternate moves that, when accepted under a strict-improvement check, replace on-policy samples in the GRPO group.

On a natural Cross-Entropy Game (an infilling task), Frost-GRPO yields qualitatively better solutions, much higher best-of-$K$ scores, and faster convergence. Frost-GRPO preserves on-policy entropy and variance where GRPO collapses, suggesting a promising route for training LLMs out-of-distribution.

This highlights the potential of tasks with differentiable rewards (in particular Cross-Entropy Games), and of their gradients, toward significantly improving the training of LLMs. 

\section*{Acknowledgements}
The authors are grateful to Mohammad Asani, Gloria Capano, Tarun Chitra, Diego Dorn, Andrew Emil, Alex Graves, Nicola Greco, Arthur Jacot, Vassilis Papadopoulos, David Pfau, Yi Sun, Alexandre Variengen, and Jérémie Wenger for many stimulating discussions and suggestions. 

\bibliographystyle{plain}
\bibliography{references}

@inproceedings{ebrahimi2018hotflip,
  title     = {{H}ot{F}lip: White-Box Adversarial Examples for Text Classification},
  author    = {Ebrahimi, Javid and Rao, Anyi and Lowd, Daniel and Dou, Dejing},
  booktitle = {Annual Meeting of the Association for Computational Linguistics},
  pages     = {31--36},
  year      = {2018},
  publisher = {Association for Computational Linguistics}
}

@inproceedings{shin2020autoprompt,
  title     = {{A}uto{P}rompt: {E}liciting {K}nowledge from {L}anguage {M}odels with {A}utomatically {G}enerated {P}rompts},
  author    = {Shin, Taylor and Razeghi, Yasaman and Logan, Robert L. and Wallace, Eric and Singh, Sameer},
  booktitle = {Conference on Empirical Methods in Natural Language Processing},
  publisher = {Association for Computational Linguistics},
  pages     = {4222--4235},
  year      = {2020}
}

@misc{zou2023universal,
  title         = {Universal and Transferable Adversarial Attacks on Aligned Language Models},
  author        = {Zou, Andy and Wang, Zifan and Carlini, Nicholas and Nasr, Milad and Kolter, J. Zico and Fredrikson, Matt},
  year          = {2023},
  eprint        = {2307.15043},
  archivePrefix = {arXiv}
}

@inproceedings{kumar2022gradient,
  title     = {Gradient-based Constrained Sampling from Language Models},
  author    = {Kumar, Sachin and Paria, Biswajit and Tsvetkov, Yulia},
  booktitle = {Conference on Empirical Methods in Natural Language Processing},
  year      = {2022},
  publisher = {Association for Computational Linguistics}
}

@inproceedings{qin2022cold,
  title     = {{COLD} Decoding: Energy-based Constrained Text Generation with Langevin Dynamics},
  author    = {Qin, Lianhui and Welleck, Sean and Khashabi, Daniel and Choi, Yejin},
  booktitle = {Advances in Neural Information Processing Systems},
  year      = {2022}
}

@inproceedings{bharadhwaj2020gradcem,
  title     = {Model-Predictive Control via Cross-Entropy and Gradient-Based Optimization},
  author    = {Bharadhwaj, Homanga and Xie, Kevin and Shkurti, Florian},
  booktitle = {Conference on Learning for Dynamics and Control},
  year      = {2020},
  series    = {Proceedings of Machine Learning Research},
  publisher = {PMLR}
}

@misc{huang2021cemgd,
  title         = {{CEM-GD}: Cross-Entropy Method with Gradient Descent Planner for Model-Based Reinforcement Learning},
  author        = {Huang, Kevin and Lale, Sahin and Rosolia, Ugo and Shi, Yuanyuan and Anandkumar, Anima},
  year          = {2021},
  eprint        = {2112.07746},
  archivePrefix = {arXiv},
}

@inproceedings{lin2020taylorgan,
  title     = {{TaylorGAN}: Neighbor-Augmented Policy Update Towards Sample-Efficient Natural Language Generation},
  author    = {Lin, Chun-Hsing and Wu, Siang-Ruei and Lee, Hung-Yi and Chen, Yun-Nung},
  booktitle = {Advances in Neural Information Processing Systems},
  year      = {2020}
}

@article{shao2024deepseekmath,
  title         = {{DeepSeekMath}: Pushing the Limits of Mathematical Reasoning in Open Language Models},
  author        = {Shao, Zhihong and Wang, Peiyi and Zhu, Qihao and Xu, Runxin and Song, Junxiao and Bi, Xiao and Zhang, Haowei and Zhang, Mingchuan and Li, Y. K. and Wu, Y. and Guo, Daya},
  journal       = {arXiv preprint arXiv:2402.03300},
  year          = {2024},
}

@inproceedings{sutton1999policygradient,
  title     = {Policy Gradient Methods for Reinforcement Learning with Function Approximation},
  author    = {Sutton, Richard S. and McAllester, David and Singh, Satinder and Mansour, Yishay},
  booktitle = {Advances in Neural Information Processing Systems},
  year      = {1999},
  publisher = {MIT Press},
}

@article{schulman2017ppo,
  title         = {Proximal Policy Optimization Algorithms},
  author        = {Schulman, John and Wolski, Filip and Dhariwal, Prafulla and Radford, Alec and Klimov, Oleg},
  journal       = {arXiv preprint arXiv:1707.06347},
  year          = {2017},
}

@article{williams1992reinforce,
  title         = {Simple Statistical Gradient-Following Algorithms for Connectionist Reinforcement Learning},
  author        = {Williams, Ronald J.},
  journal       = {Machine Learning},
  year          = {1992},
  doi           = {10.1007/BF00992696}}

@misc{hongler2025crossentropy,
  title         = {Cross-Entropy Games for Language Models: From Implicit Knowledge to General Capability Measures},
  author        = {Hongler, Clément and Emil, Andrew},
  year          = {2025},
  eprint        = {2506.06832},
  archivePrefix = {arXiv},
}

@misc{hongler2026cognitive,
  title         = {Cognitive Training for Language Models: Towards General Capabilities via Cross-Entropy Games},
  author        = {Hongler, Clément and Gabriel, Franck and Hartmann, Valentin and Renard, Arthur and Emil, Andrew},
  year          = {2026},
  eprint        = {2603.22479},
  archivePrefix = {arXiv},
}

@inproceedings{hu2024amortizing,
  title     = {Amortizing Intractable Inference in Large Language Models},
  author    = {Hu, Edward J. and Jain, Moksh and Elmoznino, Eric and Kaddar, Younesse and Lajoie, Guillaume and Bengio, Yoshua and Malkin, Nikolay},
  booktitle = {International Conference on Learning Representations},
  year      = {2024}
}

@article{das2024security,
author = {Das, Badhan Chandra and Amini, M. Hadi and Wu, Yanzhao},
title = {Security and Privacy Challenges of Large Language Models: A Survey},
year = {2025},
publisher = {Association for Computing Machinery},
doi = {10.1145/3712001},
journal = {ACM Computing Surveys},
}

@inproceedings{hatamizadeh2026rlp,
  title     = {{RLP}: Reinforcement as a Pretraining Objective},
  author    = {Hatamizadeh, Ali and Akter, Syeda Nahida and Prabhumoye, Shrimai and Kautz, Jan and Patwary, Mostofa and Shoeybi, Mohammad and Catanzaro, Bryan and Choi, Yejin},
  booktitle = {International Conference on Learning Representations},
  year      = {2026}
}

@misc{huggingface_cosmopedia,
  title        = {Cosmopedia},
  author       = {{Hugging FaceTB}},
  year         = {2024},
  note         = {Apache License 2.0. Accessed: 2026-05-07},
  howpublished = {\url{https://huggingface.co/datasets/HuggingFaceTB/cosmopedia}},

}

@misc{qwen3technicalreport,
  title         = {Qwen3 Technical Report},
  author        = {{Qwen Team}},
  year          = {2025},
  eprint        = {2505.09388},
  archivePrefix = {arXiv},
}

\newpage
\appendix

\section{Fraction of parents replaced}
\label{app:frac-replaced}

\begin{figure}[H]
\centering
\includegraphics[width=0.8\linewidth]{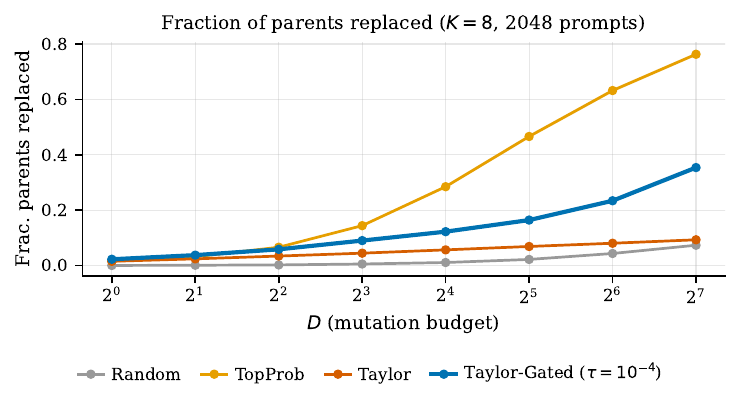}
\caption{Fraction of the $K = 8$ parents that received at least one improving mutation, as a function of the discovery budget $D \in \{1, 2, 4, 8, 16, 32, 64, 128\}$, averaged over $128$ validation prompts (shaded $\pm 1$ standard error).}
\label{fig:app1}
\end{figure}

This figure supports the structural-diversification argument in the paper. TopProb climbs from near $0$ at $D = 1$ to roughly $0.78$ at $D = 128$: at any fixed $(k, j)$ slot only one token can be the most probable one, so as $D$ grows TopProb is mechanically forced to draw from new $(k, j)$ pairs and therefore touches more and more distinct parents. Taylor and Taylor-Gated rank by the Taylor approximation, which has no such structural diversification: when one $(k, j)$ slot has a large gradient norm, the top of the Taylor ranking concentrates several replacement tokens at that single slot. As a result, Taylor-Gated tops out around $0.35$ of parents replaced and Taylor around $0.10$ even at $D = 128$. Random sits near $0$ throughout, since uniform-random mutations rarely beat their parent.

Read together with Figure~\ref{fig:sweepK} in the main text: the rules that touch fewer parents (Taylor-Gated, Taylor) win on best-of-$K$ because the few replacements they make are large; the rule that touches the most parents (TopProb) loses on best-of-$K$ because it spends its budget on many small wins.

\section{Threshold sweep on \texorpdfstring{$\tau$}{tau}}
\label{app:threshold}

\begin{figure}[H]
  \centering
  \includegraphics[width=\linewidth]{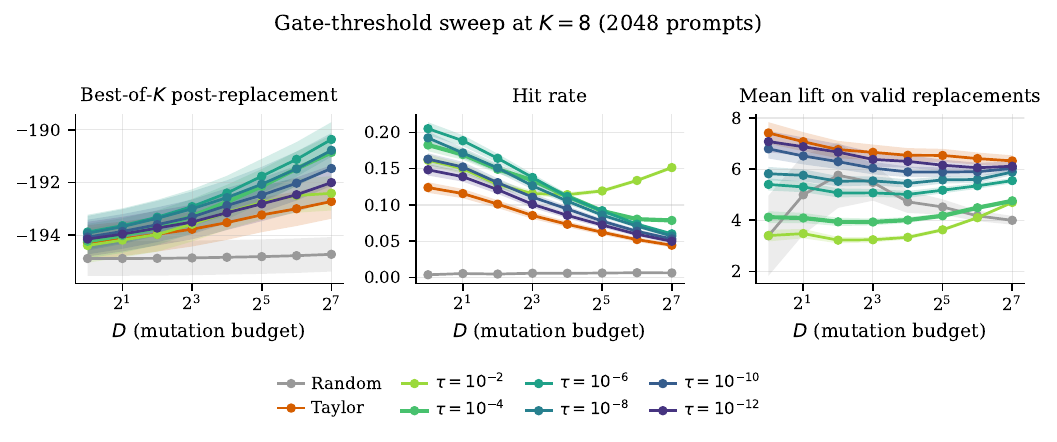}
  \caption{Threshold sweep at $K = 8$. Left: best-of-$K$ post-replacement. Middle: hit rate. Right: mean lift on valid replacements. }
  \label{fig:threshold}
\end{figure}

We use Figure~\ref{fig:threshold} to justify the choice $\tau = 10^{-4}$ adopted in this paper. On the best-of-$K$ post-replacement panel, the two least restrictive thresholds in the sweep, $\tau \in \{10^{-2}, 10^{-4}\}$, both dominate at every $D$, with the gap between them small and well within the SE bands across the full $D$ range. As $\tau$ is lowered through $10^{-6}, 10^{-8}, \dots, 10^{-12}$, the best-of-$K$ curve shifts down toward the Taylor approximation (no gate) curve, recovering the ungated behavior in the limit $\tau \to 0$. The hit-rate panel tells the same story from the candidate side: stricter gates trade the absolute number of available candidates for a higher fraction of positions whose mutation actually beats the parent, and the relationship is monotone.

We adopt $\tau = 10^{-4}$ as the slightly more conservative of the two best options. Both $\tau = 10^{-4}$ and $\tau = 10^{-2}$ are valid choices on this evidence; the rest of the paper would not change qualitatively under $\tau = 10^{-2}$.

\section{Prompt Generation}
\label{app:prompt_generation}

\paragraph{Prompt Template:} Prompts are generated from the following Python template, where
\texttt{BEGINNING}, \texttt{END}, and \texttt{MOVE\_LENGTH} are
substituted per example:

\lstset{
  basicstyle=\ttfamily\footnotesize,
  breaklines=true,
  columns=fullflexible,
  keepspaces=true,
  showstringspaces=false,
  frame=single,
  xleftmargin=0pt,
  xrightmargin=0pt,
}

\begin{lstlisting}[language=Python]
messages = [{
    "role": "user",
    "content": (
        f"A text begins with:"
        f"\n'''\n{BEGINNING}\n'''\n\n"
        f"After a gap of 24 tokens, it continues with:\n"
        f"'''\n{END}\n'''\n\n"
        f"Provide exactly {MOVE_LENGTH} tokens to bridge between the opening and the continuation, maximizing the likelihood of the full sequence. Respond with exactly {MOVE_LENGTH} tokens; only these will be used."
    ),
}]
\end{lstlisting}

\paragraph{Example of a fully instantiated prompt:}
With \texttt{BEGINNING = "Once upon a time, in a village"},
\texttt{MOVE\_LENGTH = 8}, and \texttt{END} as below:

\begin{lstlisting}[language=]
<|im_start|>user
A text begins with:
'''
Once upon a time, in a village
'''

After a gap of 24 tokens, it continues with:
'''
and learning new things every day! One day, they stumbled upon a magical forest full of vibrant colors and fascinating creatures. As they ventured deeper into the forest, they met Torty, a wise old turtle who was known to have answers to all questions.

Kiwi asked Torty, "How does our culture affect the
'''

Provide exactly 8 tokens to bridge between the opening and the continuation, maximizing the likelihood of the full sequence.
Respond with exactly 8 tokens; only these will be used.
<|im_end|>
<|im_start|>assistant
<think>

</think>
\end{lstlisting}

\newpage
% \section{Sampling Analysis}
\section{Sampling Diagnostics}
\label{app:sampling_diagnostics}

% \subsection{Aggregate diagnostics}
We report aggregate statistics over 30 prompts, computed on samples drawn at training step 8k for the canonical training.

\begin{table}[H]
\centering
\begin{tabular}{lrrrrrr}
\toprule
Method & Mean reward & Best-of-8 & Variance & Entropy & Prefix CE & Text CE \\
\midrule
Frost & $-182.87$ & $-172.77$ & $81.87$ & $0.58$ & $21.18$ & $161.69$ \\
GRPO  & $-184.00$ & $-181.53$ & $13.55$ & $0.12$ & $19.89$ & $164.11$ \\
\bottomrule
\end{tabular}
\vspace{0.7em}
\caption{Reward and diversity statistics. Prefix CE is the cross-entropy of the sampled move; Text CE is that of the continuation.}
\label{tab:appendix-c-aggregate}
\end{table}

\begin{table}[H]
    \centering
    \begin{tabular}{lr}
        \toprule Metric & Count \\
        \midrule
            Frost best-of-8 strictly exceeds GRPO best-of-8 & $25 / 30$ \\
            GRPO groups with zero score variance & $19 / 30$ \\
            Frost groups with zero score variance & $0 / 30$ \\
            GRPO best-of-8 is a repeat of the beginning & $19 / 30$ \\
            Frost wins on the remaining 11 prompts & $10 / 11$ \\
        \bottomrule
        \end{tabular}
    \vspace{0.7em}
    \caption{A zero-variance group means all eight samples received the same reward. The last row restricts to prompts where GRPO's best sample is not a copy of $x_{<k}$.}
    \label{tab:appendix-c-diagnostics}
\end{table}

 \subsection{Examples of Generated Bridges}
 \label{app:examples_of_samples}

To complement the diagnostics of the previous section, we report four representative examples illustrating different qualitative cases. The first one shows a GRPO repetition mode, where GRPO copies the beginning while Frost produces plausible bridge candidates. The next two show cases where both methods produce non-repeating continuations, but Frost obtains a better judge score. The last example is a situation where neither finds a fully satisfactory bridge, but GRPO obtains the highest judge score. 

In all the tables below, the candidates with the highest judge scores are shown in bold.

\begin{table}[H]
\small
\noindent\textbf{Beginning} ($x$): I live in Astoria, Oregon,

\noindent\textbf{End} ($z$):  and gloomy weather - but we locals like to call it "atmospheric." We get used to the constant drizzle and overcast skies after a while; they become as familiar as an old friend. Or at least that's what I thought until last year when something strange happened.\par I had always heard about how

\medskip
\begin{tabularx}{\linewidth}{@{}XX@{}}
\toprule
\textbf{Frost samples} ($y$) & \textbf{GRPO samples} ($y$) \\
\midrule
\textbf{`` a small town with a lot of rain''} \hfill \textbf{-174.088} & ``I live in Astoria, Oregon,'' \hfill -198.982 \\
`` a small town with a lot of trees'' \hfill -177.381 & ``I live in Astoria, Oregon,'' \hfill -198.982 \\
`` a quiet little town with a lot of'' \hfill -183.123 & ``I live in Astoria, Oregon,'' \hfill -198.982 \\
`` a small town on the coast, with'' \hfill -186.736 & ``I live in Astoria, Oregon,'' \hfill -198.982 \\
`` a small town in the Pacific Northwest.'' \hfill -186.821 & ``I live in Astoria, Oregon,'' \hfill -198.982 \\
`` a small town with lots of trees and'' \hfill -188.843 & ``I live in Astoria, Oregon,'' \hfill -198.982 \\
`` a place with lots of rain and glo'' \hfill -196.655 & ``I live in Astoria, Oregon,'' \hfill -198.982 \\
`` a place with a certain charm, with'' \hfill -198.33 & ``I live in Astoria, Oregon,'' \hfill -198.982 \\
\bottomrule
\end{tabularx}
\caption{GRPO collapses to repeated copies of the beginning, while Frost produces plausible bridges.}
\label{tab:bridge-candidates-13}
\end{table}

 \begin{table}[H]
 \small
 \noindent\textbf{Beginning} ($x$): Once upon a time, in a small

 \noindent\textbf{End} ($z$): . They were not ordinary plants; they had feelings and could talk! These three friends loved their town and played together every day. However, one day they noticed something strange happening around them. It was getting hotter than usual during summer, winters weren't so cold anymore, and sometimes it even rained when it shouldn't

 \medskip
 \begin{tabularx}{\linewidth}{@{}XX@{}}
 \toprule
 \textbf{Frost samples} ($y$) & \textbf{GRPO samples} ($y$) \\
 \midrule
 \textbf{`` village, there lived three very special plants''} \hfill \textbf{-157.256} & `` village, there lived three magical sunflowers'' \hfill -172.449 \\
 `` town, there were three special plants that'' \hfill -173.677 & `` village, there lived three special plants,'' \hfill -176.957 \\
 `` village, there were three special plants that'' \hfill -177.888 & `` village, there lived three special plants,'' \hfill -176.957 \\
 `` village, there were three special plants that'' \hfill -177.888 & `` village, there lived three special plants,'' \hfill -176.957 \\
 `` town, there lived three magical plants.'' \hfill -180.502 & `` village, there lived three special plants,'' \hfill -176.957 \\
 `` town, there lived three magical trees.'' \hfill -183.249 & `` village, there lived three magical plants,'' \hfill -177.973 \\
 `` village, there lived three magical plants.'' \hfill -184.756 & `` village, there lived three magical plants,'' \hfill -177.973 \\
 `` village, there lived three magical plants.'' \hfill -184.756 & `` village, there lived three magical trees,'' \hfill -183.5 \\
 \bottomrule
 \end{tabularx}
 \caption{Both methods produce non-repeating plausible continuations; Frost obtains the best score.}
 \label{tab:bridge-candidates-3}
\end{table}

 \begin{table}[H]
 \small
 \noindent\textbf{Beginning} ($x$): One sunny day, Sally the Slug and

 \noindent\textbf{End} ($z$):  it, but they wanted to find out for themselves.\par As they journeyed through the garden, Sally explained to Freddie that she produces slime to help her move around. She also uses it to protect herself from predators and to keep moist when it's dry outside.\par "But what exactly is your slime made of, Sally?" asked

 \medskip
 \begin{tabularx}{\linewidth}{@{}XX@{}}
 \toprule
 \textbf{Frost samples} ($y$) & \textbf{GRPO samples} ($y$) \\
 \midrule
 \textbf{`` Freddie the Firefly set out to explore''} \hfill \textbf{-158.525} & `` Freddie the Frog set out to explore the'' \hfill -164.893 \\
 `` Freddie the Firefly set off to explore'' \hfill -159.064 & `` Freddie the Frog set out to explore the'' \hfill -164.893 \\
 `` Freddie the Frog went on an adventure with'' \hfill -159.804 & `` Freddie the Frog went on a walk with'' \hfill -165.383 \\
 `` Freddie the Firefly went on a journey'' \hfill -164.217 & `` Freddie the Frog were walking in the garden'' \hfill -170.547 \\
 `` Freddie the Frog were playing in the garden'' \hfill -169.319 & `` Freddie the Frog were walking in the garden'' \hfill -170.547 \\
 `` Freddie the Frog were playing in the garden'' \hfill -169.319 & `` Freddie the Firefly joined her, curious'' \hfill -170.821 \\
 `` Freddie the Firefly set out on an'' \hfill -170.87 & `` Freddie the Frog decided to join her,'' \hfill -172.661 \\
 `` Freddie the Firefly were sitting in the'' \hfill -176.412 & `` Freddie the Frog decided to join her,'' \hfill -172.661 \\
 \bottomrule
 \end{tabularx}
 \caption{Both methods produce non-repeating continuations, but the best GRPO bridge leads to an awkward phrase, whereas Frost's best bridge connects the beginning cleanly with the end. }
 \label{tab:bridge-candidates-22}
 \end{table}

\begin{table}[H]
\small
\noindent\textbf{Beginning} ($x$): Once upon a time, in a village

\noindent\textbf{End} ($z$):  and learning new things every day! One day, they stumbled upon a magical forest full of vibrant colors and fascinating creatures. As they ventured deeper into the forest, they met Torty, a wise old turtle who was known to have answers to all questions.\par Kiwi asked Torty, "How does our culture affect the

\medskip
\begin{tabularx}{\linewidth}{@{}XX@{}}
\toprule
\textbf{Frost samples} ($y$) & \textbf{GRPO samples} ($y$) \\
\midrule
`` there lived a curious Kiwi who loved'' \hfill -157.225 & \textbf{``, there lived a curious kiwi named''} \hfill \textbf{-151.731} \\
`` there lived a curious Kiwi who loved'' \hfill -157.225 & \textbf{``, there lived a curious kiwi named''} \hfill \textbf{-151.731} \\
`` there lived a curious Kiwi who loved'' \hfill -157.225 & \textbf{``, there lived a curious kiwi named''} \hfill \textbf{-151.731} \\
`` there lived a curious Kiwi who loved'' \hfill -157.225 & \textbf{``, there lived a curious kiwi named''} \hfill \textbf{-151.731} \\
`` there lived a group of curious Kiwis'' \hfill -163.094 & \textbf{``, there lived a curious kiwi named''} \hfill \textbf{-151.731} \\
`` there lived a curious kiwi named Ki'' \hfill -172.501 & \textbf{``, there lived a curious kiwi named''} \hfill \textbf{-151.731} \\
`` there lived a curious kiwi named Ki'' \hfill -172.501 & \textbf{``, there lived a curious kiwi named''} \hfill \textbf{-151.731} \\
`` there was a curious kiwi named Ki'' \hfill -172.627 & ``, there lived a group of curious animals'' \hfill -159.38 \\
\bottomrule
\end{tabularx}
\caption{Neither method finds a fully satisfactory bridge, and GRPO obtains the highest judge score despite the awkward transition.}
\label{tab:bridge-candidates-0}
\end{table}

\end{document}